\def\year{2022}\relax
\documentclass[letterpaper]{article} 

\usepackage{multirow}
\usepackage{color}
\usepackage{amsfonts}
\usepackage{amsmath}
\usepackage{bm}

\newcommand{\figref}[1]{Fig.~\ref{#1}}
\newcommand{\tabref}[1]{Table~\ref{#1}}
\newcommand{\eg}[1]{\textit{e.g.,}}
\newcommand{\ie}[1]{\textit{i.e.,}}

\newcommand{\tabincell}[2]{\begin{tabular}{@{}#1@{}}#2\end{tabular}}

\usepackage{aaai22}  
\usepackage{times}  
\usepackage{helvet}  
\usepackage{courier}  
\usepackage[hyphens]{url}  
\usepackage{graphicx} 
\usepackage{natbib}  
\usepackage{caption} 
\DeclareCaptionStyle{ruled}{labelfont=normalfont,labelsep=colon,strut=off} 
\usepackage{algorithm}
\usepackage{algorithmic}

\usepackage{newfloat}
\usepackage{listings}
\floatstyle{ruled}
\newfloat{listing}{tb}{lst}{}
\floatname{listing}{Listing}
\title{Detail-Preserving Transformer for Light Field Image Super-Resolution}

\author {
	Shunzhou Wang$^{1}$\thanks{The first two authors contribute equally to this work.},
	Tianfei Zhou$^{2*}$,
	Yao Lu$^1$\thanks{Corresponding author},
	Huijun Di$^1$
}
\affiliations {
	$^1$ Beijing Key Laboratory of Intelligent Information Technology,\\ School of Computer Science and Technology, Beijing Institute of Technology, China\\
	$^2$ Computer Vision Laboratory, ETH Zurich, Switzerland\\
	\small\texttt{\{shunzhouwang, vis\_yl, ajon\}@bit.edu.cn}~~~~~ \small\texttt{tianfei.zhou@vision.ee.ethz.ch} \\
}

\begin{document}

\maketitle

\begin{abstract}
	Recently, numerous algorithms have been developed to tackle the problem of light field super-resolution (LFSR), \ie, super-resolving low-resolution light fields to gain high-resolution views. Despite delivering encouraging results, these approaches are all convolution-based, and are naturally weak in global relation modeling of sub-aperture images necessarily to characterize the inherent structure of light fields. In this paper, we put forth a novel formulation built upon Transformers, by treating LFSR as a sequence-to-sequence reconstruction task. In particular, our model regards sub-aperture images of each vertical or horizontal angular view as a sequence, and establishes long-range geometric dependencies within each sequence via a spatial-angular locally-enhanced self-attention layer, which maintains the locality of each sub-aperture image as well. Additionally, to better recover image details, we propose a detail-preserving Transformer (termed as DPT), by leveraging gradient maps of light field to guide the sequence learning. DPT consists of two branches, with each associated with a Transformer for learning from an original or gradient image sequence. The two branches are finally fused to obtain comprehensive feature representations for reconstruction. Evaluations are conducted on a number of light field datasets, including real-world scenes and synthetic data. The proposed method achieves superior performance comparing with other state-of-the-art schemes. Our code is publicly available at: \texttt{https://github.com/BITszwang/DPT}.

\end{abstract}

\section{Introduction}


Light field (LF) imaging systems offer powerful capabilities to capture the 3D information of a scene, and thus enable a variety of applications going from photo-realistic image-based rendering to vision applications such as depth sensing, refocusing, or saliency detection. However, current light field cameras naturally face a trade-off between the angular and spatial resolution, that is, a camera capturing views with a high angular sampling typically at the expense of a limited spatial resolution, and vice versa. This limits the practical applications of LF, and also motivates many efforts to study super-resolution along the angular dimension (\ie, to synthesize new views) or the spatial dimension (\ie, to increase the spatial resolution). Our work focuses on the latter one.

In comparison with the traditional 2D photograph that only records the spatial intensity of light rays, a light field additionally collects the radiance of light rays along with different directions, offering a multi-view description of the scene. A naive solution of light field super-resolution (LFSR) is to super-resolve each view independently using single image super-resolution (SISR) techniques. However, despite the recent progress of SISR, the solution is sub-optimal mainly because it neglects the intrinsic relations (\ie, angular redundancy) of different light field views, possibly resulting in angularly inconsistent reconstructions. To address this, many studies exploit complementary information captured by different sub-aperture images for high-quality reconstruction. The seminal learning-based method,~\ie,~\cite{yoon2015learning}, directly stacks 4-tuples of sub-aperture images together as an input of a SISR model. Subsequent efforts develop more advanced techniques, \eg, to explore the geometric property of a light field in multiple network branches~\cite{zhang2019residual}, to align the features of the center view and its surrounding views with deformable convolutions~\cite{wang2020light}, to encourage interactions between spatial and angular features for more informative feature extraction~\cite{wang2020spatial}, to combinatorially learn  correlations between an arbitrary pair of views for super-resolution~\cite{jin2020light}, or to gain an efficient low-rank light field representation for restoration~\cite{farrugia2019light}.


Despite the encouraging results of these approaches, they are all based on convolutional network architectures, thus inherently lack strong capabilities to model global relations among different views of light field images. In light of the ill-posed nature of the super-resolution problem, we believe that {an ideal solution should take into account as much informative knowledge in the LR input as possible}.

Motivated by the above analysis, we propose, to the best of our knowledge, the first Transformer-based model to address LFSR from a holistic perspective.  In stark contrast to existing approaches, our model treats each light field as a collection of sub-aperture image (SAI) sequences (captured along horizontal or vertical directions), and exploits self-attention to reveal the intrinsic geometric structure of each sequence. Despite the advantages of Transformers in long-range sequence modeling, it is non-trivial to apply vanilla Transformers (\eg, ViT~\cite{dosovitskiy2020image}) for super-resolution tasks, mainly because {\textbf{1)}} they represent each input image with small-size (\eg, $16\!\times\!16$ or $32\!\times\!32$) patches (\ie, tokens), which may damage fundamental structures (\eg, edges, corners or lines) in images,  and {\textbf{2)}} the vanilla fully connected self-attention design focus on establishing long-range dependencies among tokens, however, ignoring the locality in the spatial dimension. To address these problems, we introduce \textit{spatial-angular locally-enhanced self-attention (SA-LSA)}, which strengths locality using convolutions first, then promotes non-local spatial-angular dependencies of each SAI sequence.

Based on SA-LSA, we design a novel Transformer-based LFSR model, \ie, DPT, which simultaneously captures local structures within each SAI and global structures of all SAIs in the light field. Concretely, DPT  consists of  a content Transformer and a gradient Transformer to learn spatial-angular dependencies within each light field and the corresponding  gradient images, respectively. It is also equipped with a cross-attention fusion Transformer to aggregate feature representations of the two branches, from which a high-resolution light field is reconstructed.

In a nutshell, our contributions are three-fold:
{\textbf{1)}} We reformulate the problem of LFSR from a sequence-to-sequence learning perspective, which is differentiated to prior works in a sense that it fully explores non-local contextual information among all sub-aperture images, better characterizing  geometric structures of light fields;
{\textbf{2)}} We design a spatial-angular locally-enhanced self-attention layer, which, in comparison with its vanilla fully-connected counterparts, offers our Transformer a strong ability to  maintain crucial local context within light fields; {\textbf{3)}} We finally introduce DPT as a novel Transformer-based architecture, which not only mines non-local contexts from multiple views, but also preserving image details for each single view. Our DPT demonstrates the promising performance on multiple benchmarks, while maintaining similar network parameters and computational cost as existing convolution-based networks.

\section{Related Work}
\noindent\textbf{Light Field Super-Resolution.} 
 Early deep learning-based methods usually respectively learn the spatial and angular information with two independent subnetworks: one subnetwork captures the spatial information and another one learns the angular information. For example,~\cite{yoon2017light} adopted the SRCNN~\cite{dong2014learning} to separately process each SAI for spatial super-resolution and interpolated novel views for angular super-resolution with the angular super-resolution network.~\cite{yuan2018light} utilized a single image super-resolution network to enlarge the spatial resolution of each SAI, and applied the proposed EPI enhancement network to restore the geometric consistency of different SAIs. Recently, many researchers seek to simultaneously capture the spatial and angular information with a unified framework.~\cite{wang2018lfnet} built a horizontal and a vertical recurrent network to respectively super-resolve 3D LF data.~\cite{zhang2019residual} stacked the SAIs from different angular directions as inputs and sent them to a multi-branch network to capture the spatial and angular information.~\cite{wang2020light} used the deformable convolution~\cite{dai2017deformable} to align and aggregate the center-view and surrounding-view features to conduct LFSR.~\cite{wang2020spatial} developed a spatial-angular interaction network to learn the spatial-angular information from the macro-pixel image constructed with different SAIs. However, a major drawback of these approaches is that they fail to consider the long-range dependency among multiple SAIs in learning rich spatial-angular representations. To address this issue, we propose a Transformer based LFSR model, in which three proposed Transformers are leveraged to establish the non-local relationship of different SAIs for more effective representation learning.


\noindent\textbf{Transformer for Image Super-Resolution.} Attention-based models have demonstrated great successes in diverse vision tasks \cite{wang2021contextual,zhou2020matnet,mou2021cola,zhou2021group,wang2021exploring,wang2021hierarchical,zhou2021cascaded} due to their powerful representative abilities. Some attention-based methods have been recently proposed to address the  super-resolution tasks. These works can be roughly grouped into two categories: Transformer for single image super-resolution and Transformer for multiple image super-resolution. The former one mainly uses the Transformer to mine the intra-frame long-range dependency for high quality image reconstruction, \ie, IPT~\cite{chen2021pre}, SwinIR~\cite{liang2021swinir}, and ESRT~\cite{Lu2021EfficientTF}. The latter one adopts the Transformer to explore the inter-frame context information for accurate image reconstruction.~\ie~TTSR~\cite{yang2020learning} for reference-based image super-resolution and VSR~\cite{cao2021video} for video super-resolution. Motivated by these approaches, we devise the first Transformer based  architecture for LFSR.


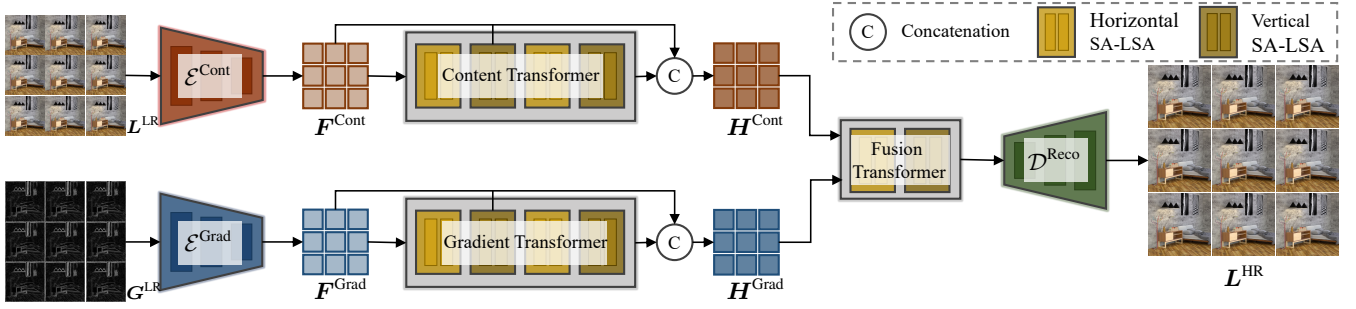
\begin{figure*}[t]
	\includegraphics[width=1\linewidth]{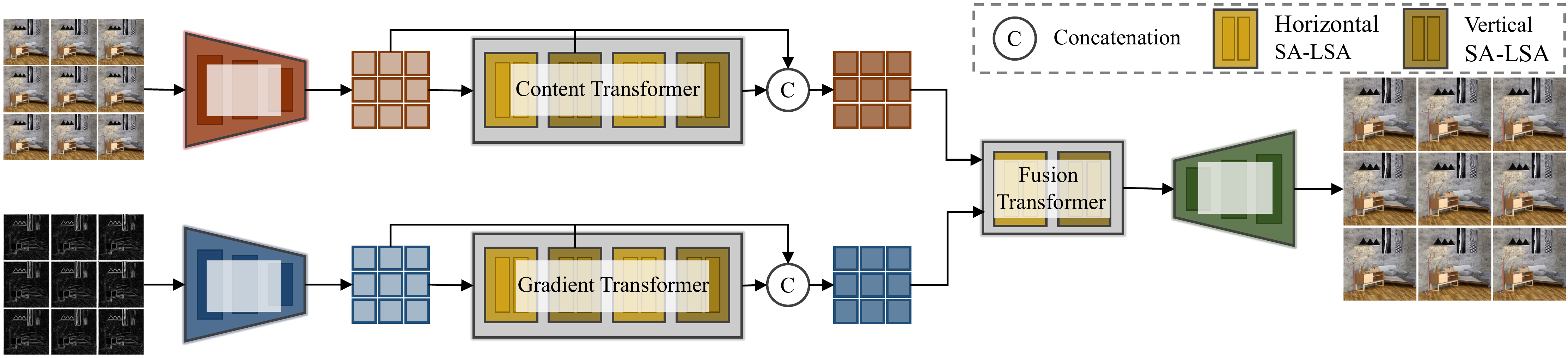}
	\put(-437,83){\small$\mathcal{E}^{\text{Cont}}$}
	\put(-437,20){\small$\mathcal{E}^{\text{Grad}}$}
	\put(-459,65){\scriptsize$\bm{L}^{\text{LR}}$}
	\put(-459,2){\scriptsize$\bm{G}^{\text{LR}}$}
	\put(-388,64){\small$\bm{F}^{\text{Cont}}$}
	\put(-388,2){\small$\bm{F}^{\text{Grad}}$}
	\put(-232,64){\small$\bm{H}^{\text{Cont}}$}
	\put(-232,2){\small$\bm{H}^{\text{Grad}}$}
	\put(-118,51){\small$\mathcal{D}^{\text{Reco}}$}
	\put(-44,7){\small$\bm{L}^{\text{HR}}$}
	\caption{\textbf{Detailed architecture of  DPT for light field image super-resolution.} Given an input light field $\bm{L}^{\text{LR}}$ and its gradient field $\bm{G}^{\text{LR}}$, our DPT leverages two separate convolutional networks (\ie, $\mathcal{E}^{\text{Cont}}$ and $\mathcal{E}^{\text{Grad}}$) for low-level feature extraction. The features are subsequently fed into a content Transformer $\mathcal{T}^{\text{Cont}}$ and a gradient Transformer $\mathcal{T}^{\text{Grad}}$, respectively, to explore global contextual information across multiple SAIs. Next, their outputs are aggregated via a cross-attention fusion Transformer $\mathcal{T}^{\text{Fuse}}$ to learn detail-preserved feature representations. Finally, the image reconstruction module $\mathcal{D}^{\text{Reco}}$ is utilized to generate the super-resolve results.}

	\label{fig:framework}
\end{figure*}

\section{Our Approach}

\noindent\textbf{Problem Formulation.} We treat the task of LFSR as a high-dimensional reconstruction problem, in which each LF is represented as a 2D angular collection of sub-aperture images (SAIs). Formally, considering an input low-resolution LF as $\bm{L}^{\text{LR}}\in\mathbb{R}^{U\times V\times H\times W}$ with angular resolution of $U\times V$ and spatial resolution of $H\times W$, LFSR aims at reconstructing a super-resolved LF $\bm{L}^{\text{HR}}\in\mathbb{R}^{U\times V\times \alpha H\times \alpha W}$ with $\alpha$ the upsampling factor. Following~\cite{yeung2018light,zhang2019residual,wang2020spatial,wang2020light}, we consider the case that SAIs distribute in a square array,
\ie, $U=V=A$, where $A$ indicates the angular resolution along the horizontal or vertical direction.



\noindent\textbf{Network Overview}. The overall architecture of DPT is shown in Fig.~\ref{fig:framework}. Given $\bm{L}^{\text{LR}}$ as the network input, we compute a gradient map for each 2D SAI, and organize them together as a gradient field  $\bm{G}^{\text{LR}}\!\in\!\mathbb{R}^{A\times A\times H\times W}$. $\bm{L}^{\text{LR}}$ and $\bm{G}^{\text{LR}}$ are separately fed into two small CNNs for convolutional feature extraction, following by two unimodal Transformers (\ie, a content Transformer and a gradient Transformer) to learn richer feature representations. To better learn relevant visual patterns among different SAIs, we treat $\bm{L}^{\text{LR}}$ (or $\bm{G}^{\text{LR}}$) as a collection of  $A$ horizontal and $A$ vertical angular sequences, and each sequence includes $A$ consecutive SAIs collected along one direction. Our content (or gradient) Transformer processes these sequences in $\bm{L}^{\text{LR}}$ (or $\bm{G}^{\text{LR}}$) one by one to avoid expensive computations as well as redundant interactions among irrelevant SAIs. Next, DPT aggregates the output features of the two Transformers via a cross-attention fusion Transformer, yielding a more comprehensive representation that is able to well preserve image details, which in final leads to better reconstruction.

In the following, we first provide a detailed description of  DPT. Then, we elaborate on  the proposed spatial-angular locally-enhanced self-attention layer, which is an essential component of our Transformer.


\subsection{Detail-Preserving Transformer (DPT)}





\noindent\textbf{Convolutional Feature Extraction.} Given the light field input $\bm{L}^{\text{LR}}$ as well as its gradient field $\bm{G}^{\text{LR}}$, two CNNs $\mathcal{E}^{\text{Cont}}$ and $\mathcal{E}^{\text{Grad}}$ are leveraged to separately extract
\begin{equation}\small\label{eq:initial}
	\begin{aligned}
		\bm{F}^{\text{Cont}} &= \mathcal{E}^{\text{Cont}}(\bm{L}^{\text{LR}})~~~\in\mathbb{R}^{A\times A\times C\times H\times W },\\
		\bm{F}^{\text{Grad}} &= \mathcal{E}^{\text{Grad}}(\bm{G}^{\text{LR}})~~~\in\mathbb{R}^{A\times A\times C\times H\times W },
	\end{aligned}
\end{equation}
per-SAI embeddings $\bm{F}^{\text{Cont}}$ and $\bm{F}^{\text{Grad}}$ at spatial resolution $H\times W$ and with embedding channel $C$.

\noindent\textbf{Content and Gradient Transformer.} The convolutional feature embeddings (\ie, $\bm{F}^{\text{Cont}}$ and $\bm{F}^{\text{Grad}}$) capture local context within each SAI independently  but lack global context across different SAIs. We use Transformers~\cite{dosovitskiy2020image} to enrich the embeddings with sequence-level context. We start by learning \textit{unimodal} contextualized representations $\bm{T}^{\text{Cont}}$ and $\bm{T}^{\text{Grad}}$ using a content Transformer $\mathcal{T}^{\text{Cont}}$ and a gradient Transformer $\mathcal{T}^{\text{Grad}}$:
\begin{equation}\small\label{eq:initial}
	\begin{aligned}
		\bm{T}^{\text{Cont}} &= \mathcal{T}^{\text{Cont}}(\bm{F}^{\text{Cont}})~~~\in\mathbb{R}^{A\times A\times C\times H\times W },\\
		\bm{T}^{\text{Grad}} &= \mathcal{T}^{\text{Grad}}(\bm{F}^{\text{Grad}})~~~\in\mathbb{R}^{A\times A\times C\times H\times W }.
	\end{aligned}
\end{equation}
Note that the two transformers share the same network structure. For simplicity, we only describe  $\mathcal{T}^{\text{Cont}}$ in the following.



The $\mathcal{T}^{\text{Cont}}$ is comprised of $K$ spatial-angular attention blocks. Each block includes two consecutive SA-LSA layers (detailed in the next section), which exploit global relations within each horizontal or vertical sequence of the input LF image, respectively.

In particular, in the $k$-th block, we treat its input as a set of horizontal (or row-wise) sequences, \ie, $\mathcal{R}\!=\!\{\bm{R}_i\!\in\!\mathbb{R}^{A\times C\times H\times W}\}_{i=1}^A$, where $\bm{R}_i$ indicates a sequence of convolutional features,  corresponding to the $i$-th row in the input. The first horizontal SA-LSA layer  $\mathcal{H}_k^1$ aims to explore the dependencies within each sequence independently. Specifically, for each horizontal sequence $\bm{R}_i$, we obtain a non-local representation $\hat{\bm{R}}_i$ as follows:
\begin{equation}\small\label{eq:conttr}
	\hat{\bm{R}}_i = \mathcal{H}_k^1 (\bm{R}_i)~~~\in\mathbb{R}^{A\times C\times H\times W}.
	\end{equation}
After processing all horizontal sequences in $\mathcal{R}$, we obtain a horizontal-enhanced content representation for the light field:
\begin{equation}\small\label{eq:row}
	\bm{R}^{\text{Cont}}_k = [\hat{\bm{R}}_1, \hat{\bm{R}}_2, \cdots, \hat{\bm{R}}_A]~~~\in\mathbb{R}^{A\times A\times C\times H\times W},
\end{equation}
where `[ ]' denotes the concatenation operation.

Next, the second vertical SA-LSA layer $\mathcal{H}_k^2$ accepts $\bm{R}^{\text{Cont}}_k$ as input, and explores the long-range relations of vertical (or column-wise) sequences, \ie, $\mathcal{C}\!=\!\{\bm{T}_i\!\in\!\mathbb{R}^{A\times C\times H\times W}\}_{i=1}^A$, where $\bm{T}_i$ indicates sequences of convolutional features,  corresponding to the $i$-th column in $\bm{R}^{\text{Cont}}_k$. Similarly, the vertical sequences in $\mathcal{C}$ are  transformed via $\mathcal{H}_k^2$ to produce a vertical-enhanced content representation $\bm{T}^{\text{Cont}}_k$:
\begin{equation}\small\label{eq:col}
	\bm{T}^{\text{Cont}}_k = [\hat{\bm{T}}_1, \hat{\bm{T}}_2, \cdots, \hat{\bm{T}}_A]~~~\in\mathbb{R}^{A\times A\times C\times H\times W},
\end{equation}
where $\hat{\bm{T}}_i\!=\!\mathcal{H}_k^2(\bm{T}_i)$ is the non-local representation of $\bm{T}_i$. Our content Transformer uses $K$ spatial-angular attention blocks to aggregate informative contextual knowledge among SAIs, eventually yielding a more comprehensive content representation $\bm{T}^{\text{Cont}}_K$.

Our gradient Transformer $\mathcal{T}^{\text{Grad}}$ performs in a similar way as the content Transformer.  It accepts
$\bm{F}^{\text{Grad}}$ as its input, and utilizes $K$ spatial-angular attention blocks to deliver a gradient representation $\bm{T}_K^{\text{Grad}}$.



\noindent\textbf{Cross-Attention Fusion Transformer.} 
While the content and gradient Transformers process each modality separately, we design a fusion Transformer to aggregate together their representations. To obtain more comprehensive \textit{unimodal} representations, we obtain the inputs of fusion Transformer by combining all intermediate features in the content or gradient Transformers:
\begin{equation}\small
	\begin{aligned}
		\!\!\!\!\!\!\bm{H}^{\text{Cont}} \!=\! [\bm{F}^{\text{Cont}}, \bm{T}_1^{\text{Cont}}, \cdots\!, \bm{T}_K^{\text{Cont}}] \in\mathbb{R}^{A\!\times\!A\!\times\!(K+1)C\!\times\!H\!\times\!W }, \\
		\!\!\!\!\!\!\bm{H}^{\text{Grad}} \!=\! [\bm{F}^{\text{Grad}}, \bm{T}_1^{\text{Grad}}, \cdots\!, \bm{T}_K^{\text{Grad}}] \in\mathbb{R}^{A\!\times\!A\!\times\!(K+1)C\!\times\!H\!\times\!W }, \\
	\end{aligned}
\end{equation}

 Different from $\mathcal{T}^{\text{Cont}}$ and $\mathcal{T}^{\text{Grad}}$ which capture non-local dependencies of tokens within the same sequence, our fusion Transformer aims to explore the relations between tokens of two sequences. In particular, for the $i$-th horizontal (or vertical) sequences in $\bm{H}^{\text{Cont}}$ and $\bm{H}^{\text{Grad}}$, \ie, $\bm{U}_i\in\mathbb{R}^{A\times (K+1)C\times H\times W}$ and $\bm{V}_i\in\mathbb{R}^{A\times (K+1)C\times H\times W}$, the $\mathcal{	T}^{\text{Fuse}}$ achieves a detail-preserved representation $\bm{Z}_i$ as:
\begin{equation}\small\label{eq:fus}
	\bm{Z} _i= \mathcal{T}^{\text{Fuse}}(\bm{U}_i, \bm{V}_i) \in\mathbb{R}^{A\times (K+1)C\times H\times W}.
\end{equation}
The $\mathcal{T}^{\text{Fuse}}$ performs cross-attention between its inputs, with query generated from $\bm{U}_i$, key and value from $\bm{V}_i$, to gather high-frequency information as a compensation of the content representation. We concatenate the outputs $\{\bm{Z}_i\}_i$ together to obtain the  fusion output $\bm{Z}\in\mathbb{R}^{A\times A\times (K+1)C\times H\times W}$. Note that our fusion Transformer only includes one spatial-angular attention block, and we see minor performance improvement when adding additional blocks.

\noindent\textbf{SAI Reconstruction.} Finally, we leverages a reconstruction module  $\mathcal{D}^{\text{Reco}}$ over $\bm{Z}$ to obtain a high-resolved LF $\bm{L}^{\text{HR}}$:
\begin{equation}\small\label{eq:reco}
	\bm{L}^{\text{HR}} = \mathcal{D}^{\text{Reco}} (\bm{Z}) ~~~\in\mathbb{R}^{A\times A\times \alpha H\times \alpha W}.
\end{equation}
Here, $\mathcal{D}^{\text{Reco}}$ is separately applied to each SAI.

\noindent{\textbf{Remark.}} Our DPT employs Transformers to enrich convolutional features $\bm{F}^{\text{Cont}}$ and $\bm{F}^{\text{Grad}}$ (Eq.~\ref{eq:initial}) into richer representations $\bm{H}^{\text{Cont}}$ and $\bm{H}^{\text{Grad}}$, respectively, which are further aggregated together via a fusion Transformer. In this manner, our network is able to collect informative non-local contexts within each SAI and across different SAIs, allowing for higher-quality reconstruction.

In addition, our Transformers process each sequence (or sequence fusion) independently rather than directly process all sequences together. This enables our model to meet hardware resource constraints, and more importantly, avoid redundant interactions among irrelevant sub-aperture images.


In Transformer architectures (\eg, ViT~\cite{dosovitskiy2020image}), fully-connected self-attention is extensively employed to explore non-local interactions among input tokens. However, the vanilla self-attention layer neglects local spatial context within each input token, which is crucial for image reconstruction. To remedy this limitation, we introduce a \textit{spatial-angular locally-enhanced self-attention layer} to offer our Transformer a strong capability in modeling locality.

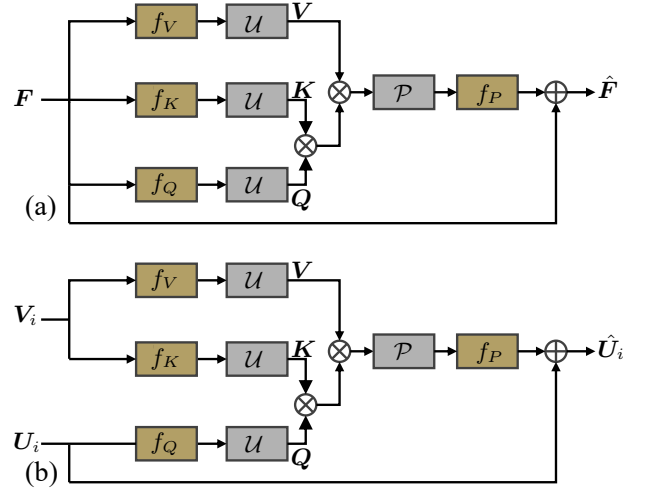
\begin{figure}[t]
	{\includegraphics[width=\linewidth]{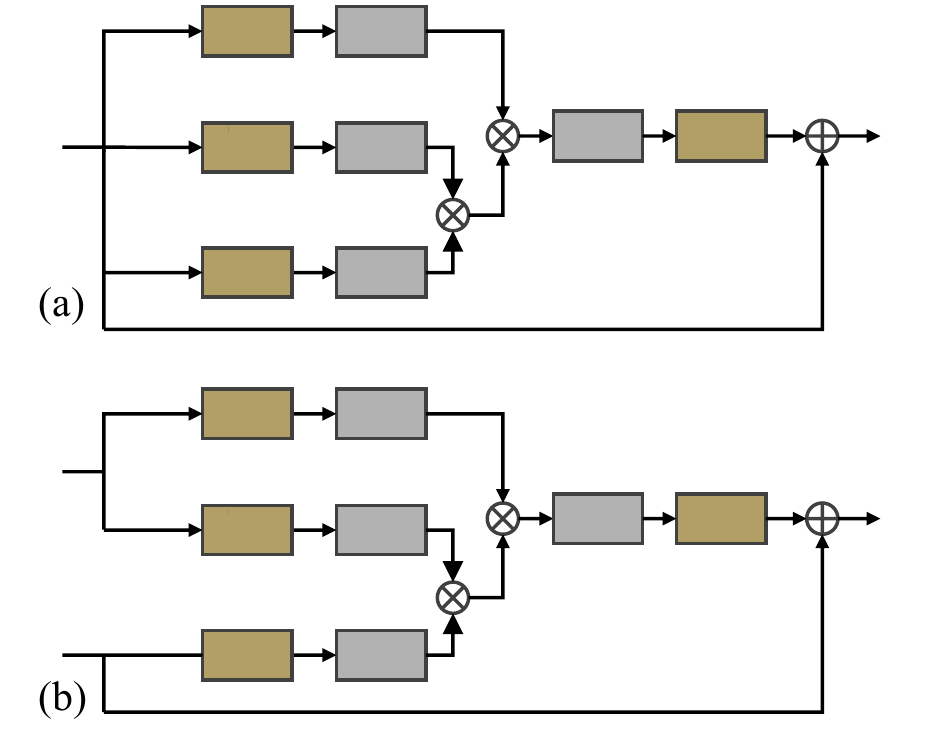}}
    \put(-234,147){\small$\bm{F}$}
    \put(-182,115){\small$f_Q$}
    \put(-182,147){\small$f_K$}
    \put(-182,176){\small$f_V$}
    \put(-146,114){\small$\mathcal{U}$}
    \put(-146,146){\small$\mathcal{U}$}
    \put(-146,175){\small$\mathcal{U}$}
    \put(-129,109){\small$\bm{Q}$}
    \put(-129,150){\small$\bm{K}$}
    \put(-129,180){\small$\bm{V}$}
    \put(-90,149){\small$\mathcal{P}$}
    \put(-59,149){\small$f_P$}
    \put(-13,150){\small$\hat{\bm{F}}$}
    \put(-234,17){\small$\bm{U}_i$}
    \put(-234,65){\small$\bm{V}_i$}
    \put(-182,17){\small$f_Q$}
    \put(-182,49){\small$f_K$}
    \put(-182,79){\small$f_V$}
    \put(-146,16){\small$\mathcal{U}$}
    \put(-146,48){\small$\mathcal{U}$}
    \put(-146,78){\small$\mathcal{U}$}
    \put(-129,11){\small$\bm{Q}$}
    \put(-129,52){\small$\bm{K}$}
    \put(-129,82){\small$\bm{V}$}
    \put(-90,51){\small$\mathcal{P}$}
    \put(-59,51){\small$f_P$}
    \put(-13,52){\small$\hat{\bm{U}_i}$}
	\caption{ Illustration of (a) the proposed SA-LSA layer (for content / gradient Transformer) as well as (b) cross-attention SA-LSA (for fusion Transformer). $\otimes$ and $\oplus$ denote the element-wise multiplication and summation operations, respectively.}
	\label{fig:salsa}
\end{figure}

\subsection{Spatial-Angular Locally-Enhanced Self-Attention}


Inspired by VSR~\cite{cao2021video}, our SA-LSA layer includes three sequential operations: spatial-angular convolutional tokenization, spatial-angular  self-attention, as well as spatial-angular  convolutional de-tokenization. Its structure is illustrated in Fig.~\ref{fig:salsa} (a).

\noindent\textbf{Spatial-Angular Convolutional Tokenization.} ViT-like architectures~\cite{dosovitskiy2020image} leverage a linear projection layer to achieve input tokens at a very early stage. In contrast, our proposed convolutional tokenization mechanism obtains the tokens in the self-attention layer, yielding a multi-stage hierarchy like CNNs. This allows our Transformer to  capture local contexts from low-level features to high-level semantic representations.

In particular, denote $\bm{F}\in\mathbb{R}^{A\times C\times H\times W}$ as the spatial-angular representation of an angular sequence, where $A$ is the length of the sequence, $C$ denotes feature dimension, and $H\times W$ is the spatial dimension. The convolutional tokenization module generates query $\bm{Q}$, key $\bm{K}$ and value $\bm{V}$ at each self-attention layer as follows:
\begin{equation}\small\label{eq:proj2}
	\bm{Q}=\mathcal{U}(f_Q(\bm{F})), ~~~\bm{K}=\mathcal{U}(f_K(\bm{F})), ~~~\bm{V}=\mathcal{U}(f_V(\bm{F})).
\end{equation}
It produces the inputs of a self-attention layer via two steps. First, the input  $\bm{F}$ is fed into three independent convolutional layers (\ie, $f_Q$, $f_K$ and $f_V$). We implement each layer with kernel size $1\times 1$.


Next, a function $\mathcal{U}$ is designed to obtain spatial-angular tokens. In particular, it extracts a collection of overlapping patches with size $H_p\times W_p$. Thus, we are able to obtain a sequence $\bm{X}\in\mathbb{R}^{n\times d}$ of $n$ tokens, and each token with a feature dimension $d=C\times H_p\times W_p$.

\noindent\textbf{Spatial-Angular Self-Attention.}
The fully-connected self-attention layer  is applied on $\bm{X}$ to explore the non-local spatial-angular relations among the tokens as follows:
\begin{equation}\small\label{eq:sa}
\bm{X}'\!=\!\texttt{Attention}(\bm{Q}, \bm{K}, \bm{V}) \!=\!  \texttt{softmax}(\bm{Q}\bm{K}^{\top}) \bm{V},
\end{equation}
where $\texttt{Attention}$ denotes a standard self-attention layer as in~\cite{vaswani2017attention,dosovitskiy2020image}.

\noindent\textbf{Spatial-Angular Convolutional De-Tokenization.}
To enable the application of convolutional tokenization in each self-attention layer, we further de-tokenize the attended feature sequence $\bm{X}'$ to fold the patches into a large feature map $\hat{\bm{F}}$ with the same dimension as $\bm{F}$:
\begin{equation}\small
	\hat{\bm{F}} = f_P(\mathcal{P}(\bm{X}'))  + \bm{F} ~~~\in\mathbb{R}^{A\times C\times H\times W},
\end{equation}
where $\mathcal{P}$ denotes the de-tokenization operation, and $f_{P}$ is a convolutional layer as Eq.~\ref{eq:proj2}. A residual layer is also used to avoid the loss of important features. Here, $\hat{\bm{F}}$ well encodes the local (in the spatial dimension) and global (in both spatial and angular dimensions) context of the input angular sequence, which could be expected to help produce better reconstruction results.

Note that in the fusion Transformer, we  improve  SA-LSA  into a cross-attention SA-LSA layer to support the computation of cross-attention between the two modalities. Fig.~\ref{fig:salsa} shows the detailed structure.

\begin{table*}[htp!]
  \centering
    \caption{Performance comparison of different methods for $\times2$ and $\times4$ SR. The best results are marked as bold.}
    \resizebox{\textwidth}{!}{
	\setlength\tabcolsep{10pt}
    \begin{tabular}{l|c||ccccc}
    \hline
    \multicolumn{1}{c|}{\multirow{1}[0]{*}{Method}} & \multicolumn{1}{c||}{\multirow{1}[0]{*}{Scale}}     & EPFL & HCInew & HCIold & INRIA & STFgantry \\
    \hline
    \hline
      Bicubic & $\times2$    & \multicolumn{1}{l}{29.50~/~0.9350} & \multicolumn{1}{l}{31.69~/~0.9335} & \multicolumn{1}{l}{37.46~/~0.9776} & \multicolumn{1}{l}{31.10~/~0.9563} & \multicolumn{1}{l}{30.82~/~0.9473} \\
    VDSR~\cite{kim2016accurate}  &  $\times2$     & \multicolumn{1}{l}{32.50~/~0.9599} & \multicolumn{1}{l}{34.37~/~0.9563} & \multicolumn{1}{l}{40.61~/~0.9867} & \multicolumn{1}{l}{34.43~/~0.9742} & \multicolumn{1}{l}{35.54~/~0.9790} \\
    EDSR~\cite{lim2017enhanced}  &  $\times2$     & \multicolumn{1}{l}{33.09~/~0.9631} & \multicolumn{1}{l}{34.83~/~0.9594} & \multicolumn{1}{l}{41.01~/~0.9875} & \multicolumn{1}{l}{34.97~/~0.9765} & \multicolumn{1}{l}{36.29~/~0.9819} \\
    RCAN~\cite{zhang2018image}  &  $\times2$     & \multicolumn{1}{l}{33.16~/~0.9635} & \multicolumn{1}{l}{34.98~/~0.9602} & \multicolumn{1}{l}{41.05~/~0.9875} & \multicolumn{1}{l}{35.01~/~0.9769} & \multicolumn{1}{l}{36.33~/~0.9825} \\
    LFBM5D~\cite{alain2018light} &  $\times2$     & \multicolumn{1}{l}{31.15~/~0.9545} & \multicolumn{1}{l}{33.72~/~0.9548} & \multicolumn{1}{l}{39.62~/~0.9854} & \multicolumn{1}{l}{32.85~/~0.9695} & \multicolumn{1}{l}{33.55~/~0.9718} \\
    GB~\cite{rossi2018geometry}    &  $\times2$     & \multicolumn{1}{l}{31.22~/~0.9591} & \multicolumn{1}{l}{35.25~/~0.9692} & \multicolumn{1}{l}{40.21~/~0.9879} & \multicolumn{1}{l}{32.76~/~0.9724} & \multicolumn{1}{l}{35.44~/~0.9835} \\
    resLF~\cite{zhang2019residual} &  $\times2$     & \multicolumn{1}{l}{32.75~/~0.9672} & \multicolumn{1}{l}{36.07~/~0.9715} & \multicolumn{1}{l}{42.61~/~0.9922} & \multicolumn{1}{l}{34.57~/~0.9784} & \multicolumn{1}{l}{36.89~/~0.9873} \\
    LFSSR~\cite{yeung2018light} &  $\times2$     & \multicolumn{1}{l}{33.69~/~0.9748} & \multicolumn{1}{l}{36.86~/~0.9753} & \multicolumn{1}{l}{43.75~/~0.9939} & \multicolumn{1}{l}{35.27~/~0.9834} & \multicolumn{1}{l}{38.07~/~0.9902} \\
    LF-InterNet~\cite{wang2020spatial} &  $\times2$     & 34.14~/~0.9761 & \multicolumn{1}{l}{37.28~/~0.9769} & \textbf{44.45}~/~\textbf{0.9945} & 35.80~/~\textbf{0.9846} & \multicolumn{1}{l}{38.72~/~0.9916} \\
    LF-DFnet~\cite{wang2020light} &  $\times2$     & {34.44}~/~\textbf{0.9766} & \textbf{37.44}~/~\textbf{0.9786} & \multicolumn{1}{l}{44.23~/~0.9943} & {36.36}~/~0.9841 & \textbf{39.61}~/~\textbf{0.9935} \\
      \textbf{DPT (Ours)} &  $\times2$  &  \textbf{34.48}~/~0.9759&{37.35~/~0.9770} & {44.31}~/~{0.9943} &  \textbf{36.40}~/~{0.9843} & {39.52}~/~{0.9928}   \\
    \hline
    Bicubic & $\times4$    & \multicolumn{1}{l}{25.14~/~0.8311} & \multicolumn{1}{l}{27.61~/~0.8507} & \multicolumn{1}{l}{32.42~/~0.9335} & \multicolumn{1}{l}{26.82~/~0.8860} & \multicolumn{1}{l}{25.93~/~0.8431} \\
    VDSR~\cite{kim2016accurate}  & $\times4$    & \multicolumn{1}{l}{27.25~/~0.8782} & \multicolumn{1}{l}{29.31~/~0.8828} & \multicolumn{1}{l}{34.81~/~0.9518} & \multicolumn{1}{l}{29.19~/~0.9208} & \multicolumn{1}{l}{28.51~/~0.9012} \\
    EDSR~\cite{lim2017enhanced}  & $\times4$    & \multicolumn{1}{l}{27.84~/~0.8858} & \multicolumn{1}{l}{29.60~/~0.8874} & \multicolumn{1}{l}{35.18~/~0.9538} & \multicolumn{1}{l}{29.66~/~0.9259} & \multicolumn{1}{l}{28.70~/~0.9075} \\
    RCAN~\cite{zhang2018image}  & $\times4$    & \multicolumn{1}{l}{27.88~/~0.8863} & \multicolumn{1}{l}{29.63~/~0.8880} & \multicolumn{1}{l}{35.20~/~0.9540} & \multicolumn{1}{l}{29.76~/~0.9273} & \multicolumn{1}{l}{28.90~/~0.9110} \\
    LFBM5D~\cite{alain2018light} & $\times4$    & \multicolumn{1}{l}{26.61~/~0.8689} & \multicolumn{1}{l}{29.13~/~0.8823} & \multicolumn{1}{l}{34.23~/~0.9510} & \multicolumn{1}{l}{28.49~/~0.9137} & \multicolumn{1}{l}{28.30~/~0.9002} \\
    GB~\cite{rossi2018geometry}    & $\times4$    & \multicolumn{1}{l}{26.02~/~0.8628} & \multicolumn{1}{l}{28.92~/~0.8842} & \multicolumn{1}{l}{33.74~/~0.9497} & \multicolumn{1}{l}{27.73~/~0.9085} & \multicolumn{1}{l}{28.11~/~0.9014} \\
    resLF~\cite{zhang2019residual} & $\times4$    & \multicolumn{1}{l}{27.46~/~0.8899} & \multicolumn{1}{l}{29.92~/~0.9011} & \multicolumn{1}{l}{36.12~/~0.9651} & \multicolumn{1}{l}{29.64~/~0.9339} & \multicolumn{1}{l}{28.99~/~0.9214} \\
    LFSSR~\cite{yeung2018light} & $\times4$    & \multicolumn{1}{l}{28.27~/~0.9080} & \multicolumn{1}{l}{30.72~/~0.9124} & \multicolumn{1}{l}{36.70~/~0.9690} & \multicolumn{1}{l}{30.31~/~0.9446} & \multicolumn{1}{l}{30.15~/~0.9385} \\
    LF-InterNet~\cite{wang2020spatial}  & $\times4$    & \multicolumn{1}{l}{28.67~/~0.9143} & \multicolumn{1}{l}{30.98~/~0.9165} & \multicolumn{1}{l}{37.11~/~0.9715} & \multicolumn{1}{l}{30.64~/~0.9486} & \multicolumn{1}{l}{30.53~/~0.9426} \\
    LF-DFnet~\cite{wang2020light} & $\times4$   & {28.77}~/~{0.9165} & \textbf{31.23}~/~\textbf{0.9196} & {37.32}~/~{0.9718} & {30.83}~/~\textbf{0.9503} & \textbf{31.15}~/~\textbf{0.9494} \\
     \textbf{DPT (Ours)}  & $\times4$  &  {\textbf{28.93}~/~\textbf{0.9167}} & {{31.19}~/~{0.9186}} & {\textbf{37.39}~/~\textbf{0.9720}} & {\textbf{30.96}~/~{0.9502}} &{{31.14}~/~{0.9487}}   \\
    \hline
    \end{tabular}}%
  \label{tab:sota}%
\end{table*}%


\subsection{Detailed Network Architecture}
\noindent\textbf{Convolutional Feature Extraction.} The convolutional modules $\mathcal{E}^{\text{Cont}}$ and $\mathcal{E}^{\text{Grad}}$ (Eq.~\ref{eq:initial}) share a similar network structure. Following~\cite{wang2020light}, each of them consists of two residual blocks and two residual atrous spatial pyramid pooling blocks organized in an intertwine manner, following by an angular alignment module without using deformable  convolution to align the convolutional features of different views. 
%


\noindent\textbf{Transformer Networks.}  In the content and gradient Transformers, we set the number of attention blocks $K$ to $2$ by default. For $\mathcal{T}^{\text{Fuse}}$, Eq.~\ref{eq:proj2} is slightly modified to enable the exploration of cross-attention by generating query from the content representation, while key and value from the gradient representation.

\noindent\textbf{Reconstruction Module.} The reconstruction module $\mathcal{D}^{\text{Reco}}$ (Eq.~\ref{eq:reco}) is implemented as a cascaded of five information multi-distillation blocks, following with a upsampling layer, which consists of a pixel shuffle operation combined with two $1\times1$ convolution layers, to generate the super-resolved SAIs~\cite{wang2020light}.




\begin{figure*}[t]
	\centerline{\includegraphics[width=1\linewidth]{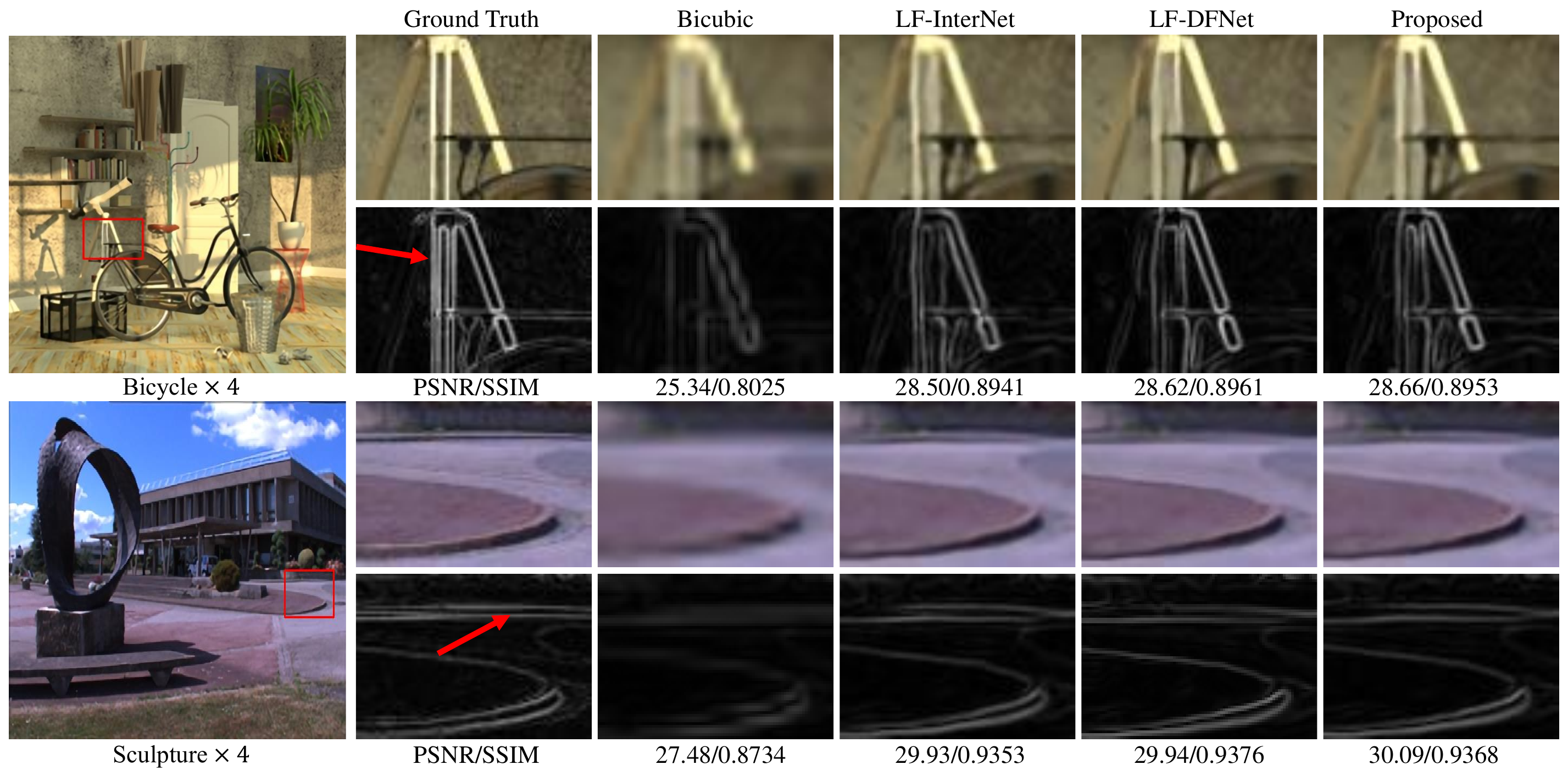}}
	\caption{Visual comparisons of \textit{Bicycle} scene from HCInew and \textit{Sculpture} scene from INRIA for $\times4$ SR. We see that our approach achieves compelling reconstruction results, with the details well preserved.}
	\label{fig:visualcomparision}
\end{figure*}

\section{Experiment}
\subsection{Datasets and Evaluation Metrics}
We conduct extensive experiments on five popular LFSR benchmarks, \ie, EPFL~\cite{rerabek2016new}, HCInew~\cite{honauer2016dataset}, HCIold~\cite{wanner2013datasets}, INRIA~\cite{le2018light}, and STFgantry~\cite{vaish2008new}. All of the light field images from the above benchmarks have a 9$\times$9 angular resolution (\ie, $U=V=A=9$). PSNR and SSIM are chosen as the evaluation metrics. For a testing dataset with $T$ scenes, we first obtain the metric values of $U\!\times\!V$ sub-aperture images, and average the summation results of $T\!\times\!U\!\times\!V$ to obtain the final metric score.
\subsection{Implementation Details}

Following~\cite{wang2020light}, we convert the light field images from the RGB space to the YCbCr space. Our model only super-resolves Y channel images, and uses the bicubic interpolation to super-resolve Cb and Cr channels images, respectively. The gradient maps are extracted from Y channel of each SAI along the spatial dimension with the function provided by~\cite{ma2020structure}. We perform $\times2$ and $\times4$ SR with $5\times5$ angular resolution on all benchmarks. In the training stage, the $64\times64$ patches are cropped from each sub-aperture image, and we apply the bicubic interpolation to generate $\times2$ and $\times4$ patches. We use random horizontal rotation, vertical rotation and $90^{\circ}$ rotation to augment the training data. Spatial and angular resolution are processed simultaneously for preserving the LF structure. The $\ell_1$ loss is used to optimize our network. We use the Adam optimizer to train our network, with a batch size of 8. The initial learning rate is set to $2\times10^{-4}$ and it will be halved every 15 epochs. We train the network for {75} epochs in total. All experiments are carried out on a single Tesla V100 GPU card.

\subsection{Comparisons with State-of-the-Art}

To evaluate the effectiveness of DPT, we compare it with the several state-of-the-art LFSR methods (~\ie~~LFBM5D~\cite{alain2018light}, GB~\cite{rossi2018geometry}, resLF~\cite{zhang2019residual}, LFSSR~\cite{yeung2018light}, LF-InterNet~\cite{wang2020spatial}, and LF-DFNet~\cite{wang2020light}) and single image super-resolution methods (~\ie~~VDSR~\cite{kim2016accurate}, EDSR~\cite{lim2017enhanced}, and RCAN~\cite{zhang2018image}) on five LFSR benchmarks. Following~\cite{wang2020light}, we  treat Bicubic upsampling as the baseline.

\noindent\textbf{Quantitative Results}. The quantitative performance comparisons are reported in \tabref{tab:sota}. As seen, DPT achieves a promising performance in comparison with other methods. Single image super-resolution methods (\ie~~VDSR~\cite{kim2016accurate}, EDSR~\cite{lim2017enhanced} and RCAN~\cite{zhang2018image}) super-resolve each view separately. DPT outperforms all of them, which can be attributed to its outstanding capability to capture the complementary information of different views.


Furthermore, compared with other CNN-based methods, \eg~ resLF~\cite{zhang2019residual}, LFSSR~\cite{yeung2018light}, LF-InterNet~\cite{wang2020spatial}, our DPT outperforms all of them for $\times4$ SR. The reason lies in that these methods are still weak in modeling global relations of different views. In contrast, our DPT can establish the global relations among different SAIs with the Transformer for efficient spatial-angular representations.

Specifically, compared with the current leading approach LF-DFNet~\cite{wang2020light}, DPT obtains superior results on EPFL~\cite{rerabek2016new}, HCIold~\cite{wanner2013datasets} and INRIA~\cite{le2018light} for $\times4$ SR. The reason is that, LF-DFNet only models the {short-range} dependencies between each side-view SAI and the center-view SAI, while our DPT explores {long-range} relations among all SAIs.

\begin{table}[t]
	\centering
	\caption{Detailed comparisons of LFSR methods
		in terms of number of parameters, {FLOPs}, and reconstruction accuracy  on the EPFL dataset for $\times2$ and $\times4$ SR.}
	\resizebox{0.49\textwidth}{!}{
		\setlength\tabcolsep{8pt}
		\begin{tabular}{l|c||cc}
			\hline
			Method & Scale & \multicolumn{1}{c}{\# Param (M) / FLOPs (G)} & \multicolumn{1}{c}{PSNR / SSIM} \\
			\hline
			\hline
			resLF & $\times$2    & 6.35~/~37.06 &   32.75~/~0.9672      \\
			LFSSR & $\times$2    & 0.81~/~25.70 &    33.69~/~0.9748     \\
			LF-InterNet & $\times$2    & 4.80~/~47.46 &   34.14~/~0.9761      \\
			LF-DFNet & $\times$2    & 3.94~/~57.22 &    34.44~/~\textbf{0.9766}     \\
			\textbf{DPT (Ours)} & $\times$2    & 3.73~/~57.44      &   \textbf{34.48}~/~{0.9759}      \\
			\hline
			resLF & $\times$4    & 6.79~/~39.70 &    27.46~/~0.8899     \\
			LFSSR & $\times$4    & 1.61~/~128.44 &   28.27~/~0.9080      \\
			LF-InterNet & $\times$4    & 5.23~/~50.10 & 28.67~/~0.9143        \\
			LF-DFNet & $\times$4    & 3.99~/~57.31 &    28.77~/~0.9165     \\
			\textbf{DPT (Ours)}  & $\times$4    &  3.78~/~58.64     & \textbf{28.93}~/~\textbf{0.9167}        \\
			\hline
	\end{tabular}}%
	\label{tab:efficience}%
\end{table}%

\begin{table*}[t!]
	\centering
	\caption{\textbf{Ablation study} of DPT on the EPFL dataset for $\times 4$ SR.}
	\resizebox{0.9\textwidth}{!}{
		\setlength\tabcolsep{8pt}
		\begin{tabular}{r|cc|ccc||c|c}
			\hline
			\multirow{2}{*}{\#} & \multirow{2}{*}{\tabincell{c}{Content\\Transformer}}  & \multirow{2}{*}{\tabincell{c}{Gradient\\Transformer}}&  \multicolumn{3}{c||}{Fusion Mechanism} & \multirow{2}{*}{\# Param (M)}&\multirow{2}{*}{PSNR / SSIM}\\
			\cline{4-6}
			
			&  &      &
			Sum  &Transformer (image)& Transformer (sequence)  &   \\
			
			\hline
			1 &      &       &       &       &       &            3.99&28.77~/~0.9165\\
			
			2 & \checkmark     &                  &       &       &       &2.62&28.77~/~{0.9142}\\
			3& \checkmark     & \checkmark           &  \checkmark &            &       &3.72&{28.86}~/~{0.9153}\\
			
			4 & \checkmark     & \checkmark        &              &  \checkmark &       &  3.75&28.81~/~0.9149\\
			5 &       &          &       &       &          \checkmark    & 3.83&{28.72}~/~{0.9139}\\
			
			6 & \checkmark     & \checkmark        &            &   & \checkmark      &3.91&{28.89}~/~{0.9157}  \\
			7   & \checkmark     & \checkmark         &             &       &  \checkmark     &3.78&{\textbf{28.93}~/~\textbf{0.9167}} \\
			\hline
	\end{tabular}}%
	\label{tab:ablation}%
\end{table*}%

\noindent\textbf{Qualitative Results}. \figref{fig:visualcomparision} depicts some representative visual results of different approaches for $\times4$ SR. As seen, LF-InterNet~\cite{wang2020spatial} and LF-DFNet~\cite{wang2020light} produce distorted results for the structures of the telescope stand in the \textit{Bicycle} scene and the flowerbed edge in the \textit{Sculpture} scene. In contrast, our DPT yields much better results, with all above regions being well preserved.

\noindent\textbf{Computational Analysis}. Table~\ref{tab:efficience} provides a detailed analysis of the LFSR models in terms of parameters, FLOPs, and reconstruct accuracy on EPFL. Following~\cite{wang2020light}, we set the size of the input LF image   as $5\times5\times32\times32$ for the computation of the FLOPs. {As can be seen,} DPT shows the best trade-off between reconstruction accuracy and model efficiency. For $\times4$ SR, DPT has fewer parameters while better accuracy, in comparison with LF-DFNet~\cite{wang2020light}. This result further confirms the effectiveness of DPT, not only in better performance but also its efficiency.


\subsection{Ablation Study}
To gain more insights into our model, we conduct a set of ablative experiments on the EPFL dataset for $\times4$ SR. The results are reported in Table~\ref{tab:ablation}. We show in the $1$st row the performance of baseline model (\ie, LF-DFNet~\cite{wang2020light}), and the $7$th row the results of our full model.


\noindent\textbf{Content Transformer.} We first investigate the effect of  the  content Transformer by constructing a network which is implemented  with the convolution feature extraction module $\mathcal{E}^{\text{Cont}}$, content Transformer $\mathcal{T}^{\text{Cont}}$ and the image reconstruction module $\mathcal{D}^{\text{Reco}}$. Its results are given in the 2nd row. We can see that the content Transformer itself can  lead to a similar performance as the baseline model, however, our content Transformer has fewer parameters. This confirms that global relations among different views brought by the content Transformer are benefical to LFSR.

\noindent\textbf{Gradient Transformer.} Furthermore, we combine the gradient Transformer into the content Transformer, with the results being shown in the $3$rd row. As we can see,  
by introducing the gradient Transformer, the performance of the content Transformer (the 2nd row) improves by 0.09dB in terms of PSNR and 0.0011 improvement in terms of SSIM, respectively. Moreover, the models with dual transformers (the 4th, the 6th and the 7th rows) outperform the model with only a content Transformer (the 2nd row), which further confirms the effectiveness of gradient Transformer. Finally, we replace the content Transformer and gradient Transformer of DPT with residual blocks while maintaining the network parameters almost unchanged. Its results are given in the 5th row. The dual branches in DPT are equivalent to two convolutional neural networks for feature extraction. As reported in \tabref{tab:ablation}, DPT has a 0.21 dB PSNR drop, demonstrating the effectiveness of the proposed Transformers.

%
%

\noindent\textbf{Impact of Fusion Mechanism.} To explore the effect of fusion mechanism for content and gradient feature aggregation, we construct two model variants, which generates the detailed-preserved representation $\bm{Z}_i$ in Eq.~\ref{eq:fus}  by the element-wise summation (the 3rd row) and the Transformer for a single view non-local dependencies exploration (the 4th row), respectively. As can be observed, our cross-attention fusion Transformer (the 7th row) brings a promising performance improvement over the results of 3rd row and the 4th row in PSNR and SSIM, which is attributed to its effectiveness for the complementary non-local information exploration of the content and gradient features.

\noindent\textbf{Efficacy of SA-LSA.} We study the effect of the SA-LSA layers by replacing them with the vanilla fully-connected self-attention layers in DPT (the 6th row). As seen, the model with SA-LSA layers (the 7th row) obtains a better performance with fewer parameters, proving the effectiveness of local spatial context for SAI reconstruction.

\noindent\textbf{Number of Attention Blocks $K$.} At last, we study the performance of DPT with respect to the number of spatial-angular attention blocks in the content Transformer (or the graidient Transformer).~\tabref{tab:numK} reports the comparison results. As seen, DPT achieves the best results at $K\!=\!2$. Thus, we choose $K\!=\!2$ as the default number of spatial-angular attention blocks in DPT.

\begin{table}[t]
	\centering
	\caption{Performance and model size comparisons with different numbers of spatial-angular attention block on the EPFL dataset for $\times4$ SR.}
	\resizebox{0.48\textwidth}{!}{
		\setlength\tabcolsep{12pt}
	\begin{tabular}{c||cccc}
		\hline
		$K$    & 1     & 2     & 3 &4\\
		\hline
		PSNR (dB)  & 28.69      & \textbf{28.93}      & 28.78 & 28.74 \\
		\# Param (M) &   2.89    &   3.78    & 5.00 & 6.56\\
		\hline
	\end{tabular}}%
	\label{tab:numK}%
\end{table}%

\section{Conclusion}

This work proposes a Detail Preserving Transformer (DPT), as the first application of Transformer  for LFSR. Instead of leveraging a vanilla fully-connected self-attention layer, we develop a spatial-angular locally-enhanced self-attention layer (SA-LSA) to promote non-local spatial-angular dependencies of each sub-aperture image sequence. Based on SA-LSA, we leverage a content Transformer and a
gradient Transformer to learn spatial-angular content and gradient representations, respectively. The comprehensive spatial-angular representations are further processed by a cross-attention fusion Transformer to aggregate the output of the two Transformers, from which a high-resolution light field is reconstructed. We compare the proposed network with other state-of-the-art methods over five commonly-used benchmarks, and the experimental results demonstrate that it achieves favorable performance against other competitors. 


\small
\bibliography{ref}

\end{document}